\documentclass[10pt,twocolumn,letterpaper]{article}
\usepackage{subfig}
\usepackage{cvpr}
\usepackage{times}
\usepackage{epsfig}
\usepackage{graphicx}
\usepackage{amsmath}
\usepackage{amssymb}

\usepackage[breaklinks=true,bookmarks=false]{hyperref}

\cvprfinalcopy 

\def\cvprPaperID{****} 

\begin{document}

\title{Diversify and Match: A Domain Adaptive Representation Learning Paradigm for Object Detection}

\author{Taekyung Kim \quad Minki Jeong \quad Seunghyeon Kim \quad Seokeon Choi \quad Changick Kim\\
Korea Advanced Institute of Science and Technology, Daejeon, Korea\\
{\tt\small $\{$tkkim93, rhm033, seunghyeonkim, seokeon, changick$\}$@kaist.ac.kr}
}

\maketitle

\begin{abstract}
   We introduce a novel unsupervised domain adaptation approach for object detection.
   We aim to alleviate the imperfect translation problem of pixel-level adaptations, and the source-biased discriminativity problem of feature-level adaptations simultaneously.
   Our approach is composed of two stages, \ie, Domain Diversification (DD) and Multi-domain-invariant Representation Learning (MRL).
   At the DD stage, we diversify the distribution of the labeled data by generating various distinctive shifted domains from the source domain.
   At the MRL stage, we apply adversarial learning with a multi-domain discriminator to encourage feature to be indistinguishable among the domains.
   DD addresses the source-biased discriminativity, while MRL mitigates the imperfect image translation.
   We construct a structured domain adaptation framework for our learning paradigm and introduce a practical way of DD for implementation.
   Our method outperforms the state-of-the-art methods by a large margin of 3$\%$ $\sim$ 12$\%$ in terms of mean average precision (mAP) on various datasets.
\end{abstract}

\section{Introduction}
Object detection is a fundamental problem in computer vision as well as machine learning.
With the recent advances of the convolutional neural networks (CNNs), CNN-based methods~\cite{RCNN,FASTR-CNN,FasterR-CNN, SSD, YOLO, RetinaNet,R-FCN,RefineDet,RFBNet} have achieved significant progress in object detection based on fine benchmarks~\cite{PascalVoc,MSCOCO,openimages}. 
Despite the promising results, all of these object detectors suffer from the degenerative problem when applied beyond these benchmarks.
Building datasets for a specific application can temporarily resolve this problem, nevertheless, the time and monetary costs incurred when manually annotating such datasets are not negligible~\cite{LabelingIssHard1, LabelingIssHard2}.
Moreover, since the intrinsic causes of the degenerative problem have been avoided instead of resolved, another generalization issue arises when extending the same application to different environments.
To address this issue, an unsupervised domain adaptation method for object detection~\cite{DAFRCNN} was recently proposed.

\begin{figure}[t]
\begin{center}
  \includegraphics[width=1.0\linewidth]{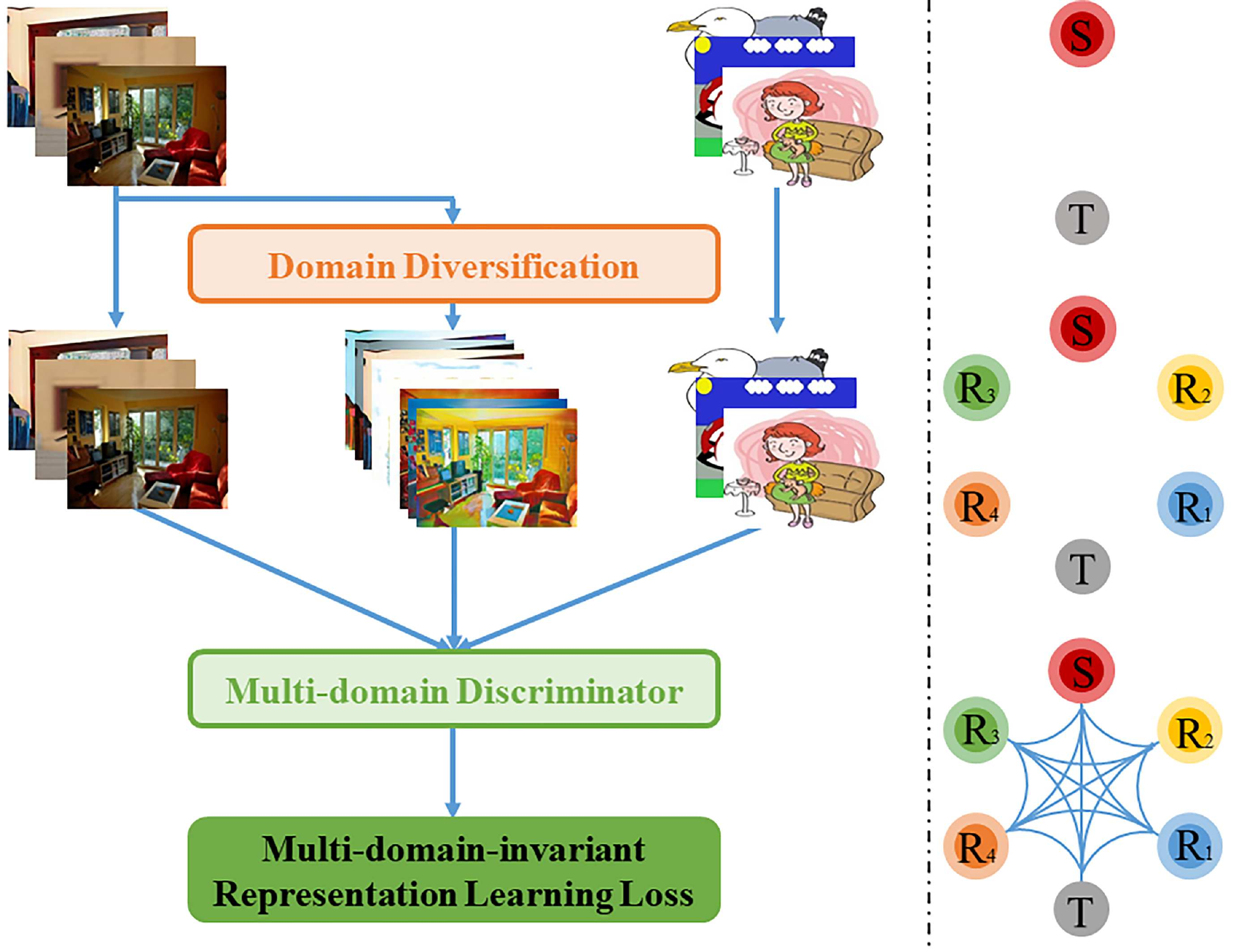}
\end{center}
  \caption{Overview of our learning paradigm. We illustrate a conceptual diagram of the distributions of the domains on the right side. S and T represent for the source and the target domain, respectively, and each $R_i$ represents the $i$th diversified domain.}
\label{fig:Intro}
\end{figure}

\begin{figure}[t]
\begin{center}
  \subfloat[\footnotesize Source domain]{
  	\begin{minipage}{0.32\linewidth}
  	\centering
    \includegraphics[width=1.0\linewidth]{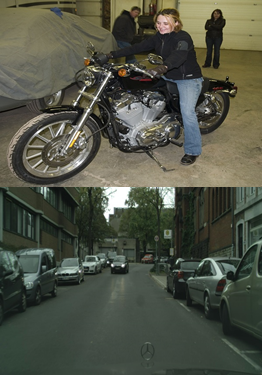}
  \end{minipage}
  }%
  \subfloat[\footnotesize Target domain]{
  \begin{minipage}{0.32\linewidth}
  \centering
      \includegraphics[width=1.0\linewidth]{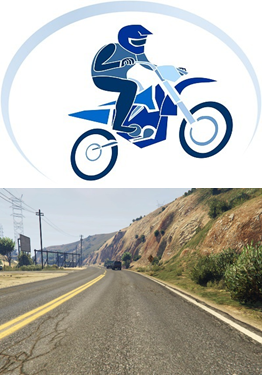}
  \end{minipage}
  }%
  \subfloat[\footnotesize Translated domain]{
  \begin{minipage}{0.32\linewidth}
  \centering
      \includegraphics[width=1.0\linewidth]{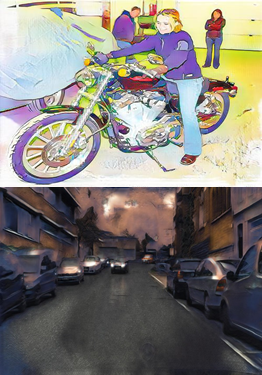}
  \end{minipage}
  }\vskip -2pt%
\end{center}\vskip -3pt
  \caption{Examples of the imperfect image translation. The first and second rows visualize examples of the translated image from the real-world to artistic media and between urban scenes, respectively.}
\label{fig:fig2}
\vskip -3pt
\end{figure}

Unsupervised domain adaptation has been studied to address the degeneration issue between related domains, which is closely related to the aforementioned degenerative problem.
With the rise of the deep neural networks, recent unsupervised deep domain adaptation methods~\cite{DeepAlign1_DAN, DeepAlign2_GRL,DeepAlign3_DCORAL, DeepAlign4_AutoDial, pixel1_sbadagan, pix3_pixlevel, seg1_CyCADA} are mainly based on feature-level adaptation and pixel-level adaptation.
Feature-level adaptation methods~\cite{DeepAlign1_DAN,DeepAlign2_GRL,DeepAlign3_DCORAL,DeepAlign4_AutoDial} align the distributions of the source and the target domain toward a cross-domain feature space.
These approaches expect the model supervised by the labeled source domain to infer on the target domain effectively.
However, the supervision of the inference layer mainly relies on the source domain only in the feature-level adaptation methods.
Thus, the feature extractor of the model is enforced to manufacture the features in a way discriminative for the source domain data, which is not suitable for the target domain.
Moreover, since the object detection data is interwoven with the instances of interest and the relatively unimportant background, it is further hard for the source-biased feature extractor to extract discriminative features for the target domain instances.
Thus, object detectors adapted at the feature-level are at risk of the source-biased discriminativity and it can leads to false recognition on the target domain.
On the other hand, pixel-level adaptation methods~\cite{pixel1_sbadagan, pix3_pixlevel, seg1_CyCADA} focus on visual appearance translation toward the opposite domain.
The model can then take advantage of the information from the translated source images~\cite{seg1_CyCADA, pix3_pixlevel} or infer pseudo label of the translated target images~\cite{Inoue_2018_CVPR}.
Most existing pixel-level adaptation methods~\cite{pixel1_sbadagan, pix3_pixlevel, seg1_CyCADA} are based on the assumption that the image translator can perfectly convert one domain to the opposite domain such that the translated images can be regarded as those from the opposite domain.
However, these methods reveal imperfect translation in many adaptation cases since the performance of the translator heavily depends on the appearance gap between the source and the target domain, as shown in Fig.~\ref{fig:fig2}. 
Regarding these incompletely translated source images as from the target domain can cause new domain discrepancy issue.

To tackle the aforementioned limitations, we introduce a novel domain adaptation paradigm for object detection.
Our learning paradigm consists of Domain Diversification (DD) and Multi-domain-invariant Representation Learning (MRL), as shown in Fig.~\ref{fig:Intro}.
Unlike most existing domain adaptation methods, DD intentionally causes several distinctive shifted domains from the source domain to enrich the distribution of the labeled data.
On the other hand, MRL boosts the domain invariance of the features by unifying the scattered domains.
Using the aforementioned approaches, we propose a universal domain adaptation framework for object detection.
Our framework trains domain-invariant object detection layers with diversified annotated data while simultaneously encouraging dispersed domains toward a common feature space. 
To demonstrate the effectiveness of our method, we conduct extensive experiments on Real-world Datasets~\cite{PascalVoc}, Artistic Media Datasets~\cite{Inoue_2018_CVPR}, and Urban Scene Datasets~\cite{CityscapesDataset, Foggy} based on Faster R-CNN.
Our framework achieves state-of-the-art performance on various datasets.

In summary, we have three contributions in our paper:
\begin{itemize} 
    \item We propose a novel learning paradigm for unsupervised domain adaptation. Our learning approach addresses the source-biased discriminativity issue and the imperfect translation issue.
    \item We structurize our learning paradigm by integrating DD and MRL in the form of a framework. 
    \item We conduct extensive experiments to validate the effectiveness of our method on various datasets.
    Our method outperforms the state-of-the-art methods with a large margin by 3$\%$ $\sim$ 12$\%$ mAP.
\end{itemize}

{
\footnotesize
\begin{figure*}
\begin{center}

\subfloat[\footnotesize Feature-level adaptation]{
  \begin{minipage}{0.18\linewidth}
  \centering
      \includegraphics[width=1.0\linewidth]{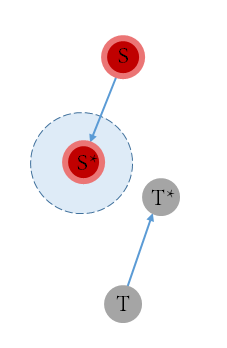}
  \end{minipage}
  }
  \subfloat[\footnotesize Pixel-level adaptation]{
  \begin{minipage}{0.18\linewidth}
  \centering
      \includegraphics[width=1.0\linewidth]{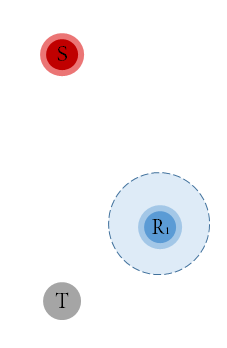}
  \end{minipage}
  }
  \subfloat[\footnotesize Domain Diversification]{
  \begin{minipage}{0.30\linewidth}
  \centering
      \includegraphics[width=1.0\linewidth]{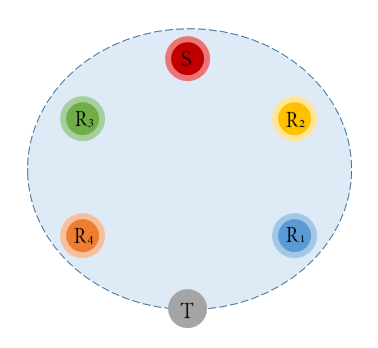}
  \end{minipage}
  }
  \subfloat[\footnotesize MRL with Domain Diversification]{
  \begin{minipage}{0.285\linewidth}
  \centering
      \includegraphics[width=1.0\linewidth]{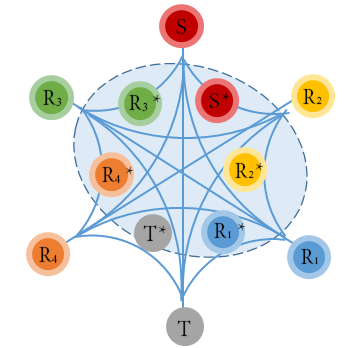}
  \end{minipage}
  }
\end{center}\vskip -3pt
  \caption{Comparison of distribution transformation by different domain adaptation methods. MRL refers to Multi-domain-invariant Representation Learning. $S$ and $T$ denote the source domain and the target domain, respectively. $R_1$, $R_2$, $R_3$, and $R_4$ are shifted domains of the source domain. The arrows indicate the feature-level adaptation trends. The domains with asterisks denote the results of feature-level adaptation. The domains with a boundary imply that the object detection network is supervised by these domains.}
\label{fig:MethodDiagram}
\vskip -3pt
\end{figure*}
}


\section{Related work}
\subsection{CNN-based Object Detection}
Traditional methods~\cite{Traditional1,Traditional2} use a sliding window framework with handcrafted features and shallow inference models.
With rise of the convolutional neural networks, R-CNN~\cite{RCNN} obtains a promising result with a selective search algorithm and classification through the CNN features.
Fast R-CNN~\cite{FASTR-CNN} reduces the bottleneck of R-CNN by sharing features among regions in the same image.
Faster R-CNN~\cite{FasterR-CNN} adopts a fully convolutional network called a Region Proposal Network (RPN) to mitigate another bottleneck caused by the selective search algorithm. 
YOLO~\cite{YOLO} achieves significant improvement in the inference speed using a single-staged network.
SSD~\cite{SSD} uses multi-scale features to enhance the relatively low accuracy of YOLO.
RetinaNet~\cite{RetinaNet} further improves the performance of single-staged object detectors using the focal loss to reduces the performance degradation caused by easy negative examples.
While these methods push the limit on the large-scale datasets with rich annotations, generalization errors which arises during their application have not been investigated thus far.

\subsection{Unsupervised Domain Adaptation}

Domain adaptation has been studied intensely in relation to the image classification task~\cite{Reweight1, Reweight2}.
Traditional methods focus on reducing domain discrepancy through instance re-weighting~\cite{Reweight1, Reweight2, Reweight3} and shallow feature alignment strategies~\cite{ShallowAlign1, ShallowAlign2}.
With the success of deep learning scheme, early deep domain adaptation mainly arises into Maximum Mean Discrepancy (MMD) minimization~\cite{DeepAlign1_DAN, DeepAlign3_DCORAL, DeepAlign4_AutoDial} or feature confusion through adversarial learning~\cite{DeepAlign2_GRL}.
Recently, as the image-to-image translation has become highlighted with promising results~\cite{ImTrans1_pix2pix, ImTrans2_DiscoGAN, ImTrans3_CoGAN, CycleGAN} through Generative Adversarial Networks (GANs)~\cite{GAN}, pixel-level adaptation methods~\cite{pixel1_sbadagan, pix2_dubplex, pix3_pixlevel} have been developed to address the domain shift issue by translating source domain images into the target style.
As unsupervised domain adaptation attracted considerable interest with its effectiveness, recent works~\cite{seg1_CyCADA,seg2_curriculum,seg3_crossdomain,seg4_ROAD,seg5_FCNsWild, seg6_StructuredOutput, seg7_ConditionalGAN, seg8_FullyConv} have been attempted to address the generalization issue in the semantic segmentation task.

Despite the recent success of unsupervised domain adaptation in various computer vision tasks, unsupervised domain adaptation for the object detection task has not been explored so far except few pioneers~\cite{Inoue_2018_CVPR,DAFRCNN}. 
Inoue et al.~\cite{Inoue_2018_CVPR} adopt a conventional unsupervised pixel-level domain adaptation method as part of a two-staged weakly supervised domain adaptation framework.
Chen et al.~\cite{DAFRCNN} align distributions of the source and the target domain at the image level and instance level to address various causes of the domain shift separately.
While these methods address the problem of degeneracy without considering the limitations of existing domain adaptation approaches, we aim to mitigate these issues through a two-step learning paradigm.



\section{Methods}
We propose a novel learning paradigm to alleviate the source-biased discriminativity in feature-level adaptation and the imperfect translation in pixel-level adaptation.
We start by explaining the two stages of our method, Domain Diversification and Multi-domain-invariant Representation Learning.
Then, a universal domain adaptation framework for object detection is introduced.
Figure~\ref{fig:MethodDiagram} shows conceptual description of feature-level adaptation, pixel-level adaptation, and our method.

\begin{figure}[t]
\vskip -7pt
\begin{center}
  \subfloat[\footnotesize Given image]{
  \begin{minipage}{0.23\linewidth}
  \centering
      \includegraphics[width=1.0\linewidth]{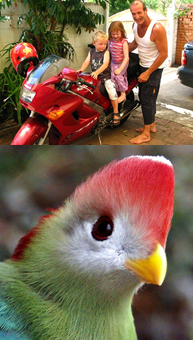}
  \end{minipage}
  }
  \subfloat[\footnotesize Images with appearance shift]{
  \begin{minipage}{0.69\linewidth}
  \centering
      \includegraphics[width=1.0\linewidth]{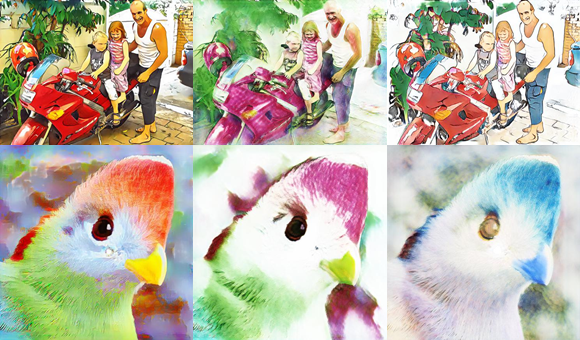}
  \end{minipage}}
\end{center}\vskip -3pt
  \caption{Examples of variously shifted images for given images.}
\label{fig:ExShift}
\vskip -3pt
\end{figure}
\subsection{Domain Diversification} \label{sec:DD}
Without loss of generality, we assume that there exist numerous possibilities of shifted domains that preserve the corresponding semantic information of the source domain but appear in different ways.
For instance, as shown in Fig.~\ref{fig:ExShift}, we can easily conceive of various visually shifted images from a given image regardless of the existence of a feasible image translator.
Along the same line, numerous variations of image translators can achieve considerable domain shift from the given source domain, which we call domain shifters.
Domain Diversification (DD) is a method which diversifies the source domain by intentionally generating distinctive domain discrepancy through these domain shifters.
The diversified distribution of the labeled data encourages the model to infer among data with large intra-class variance discriminatively.
Thus, the model is enforced to extract semantic features that are not biased to a particular domain.
This allows the model to extract unbiased semantic features from the target domain, which is more discriminative than the source-biased features.
With the better discriminativity of target domain features, we can assimilate the domains with less feature collapse, resulting in more desirable adaptation.

Among the plenteous possibilities of domain shifters, inspired by the limitation of pixel-level adaptation, we practically realize the possibilities using the imperfections of the image translation.
\begin{figure*}
\begin{center}
\includegraphics[width=1.0\linewidth]{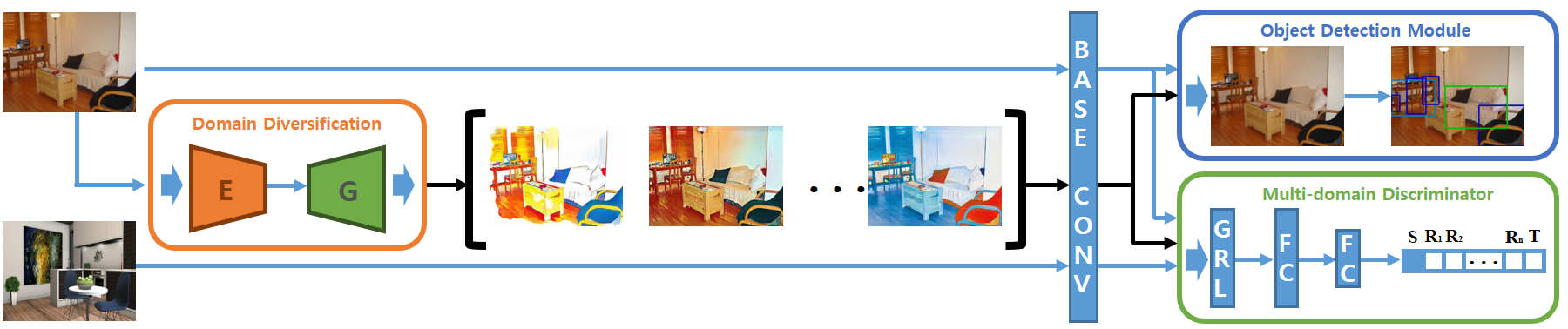}
\end{center}
  \caption{The architecture of our domain adaptation framework for object detection. Our framework is built on the object detection network.}
\label{fig:NetArch}
\end{figure*}
Let us denote a source domain sample as $x^s$ and a target domain sample as $x^t$ with domain distributions $p_{s}$ and $p_{t}$, respectively.
In general, image translation methods aim to train a generator $G$ by optimizing the translated image $G(x^s)$ to which appears to be sampled from the target domain.
However, since the generator network has high enough capacity for various translations, the adversarial loss alone cannot guarantee the conversion of a given $x^s$ to the desired target image.
To redeem this instability, image translation methods add constraints to the objective function $L_{\text{im}}$ to reduce the possibility of the undesirable generators:
\begin{equation}
    \begin{aligned}[b]
        L_{\text{im}}(G,D,M) &= L_{\text{GAN}}(G,D) + \alpha
        L_{\text{con}}(G,M),\\
    \end{aligned}
\label{eqn1}
\end{equation}
\begin{equation}
    \begin{aligned}[b]    
        L_{\text{GAN}}(G,D)& = \mathbb{E}_{x^t\sim p_{\text{t}}(x^t)}[logD(x^t)]\\
        &+\mathbb{E}_{x^s\sim p_{\text{s}}(x^s)}[log(1-D(G(x^s)))],
    \end{aligned}
\label{eqn2}
\end{equation}
where $D$ is the discriminator for adversarial learning, $L_{con}(G,M)$ is the constraint loss with a possibly existing additional module $M$ and $\alpha$ is a weight that balances the two losses.
Here, the additional module implies a supplemental network necessary for a sophisticated constraint.

In this basic setting, we observe that varying the learning trend with alternative constraints causes the generator $G$ to diversify the appearance of the translated images.
Based on this observation, we apply several variants of constraints to achieve distinct domain shifters.
The objective function for the domain shifter can be written as:
\begin{equation}
    \begin{aligned}[b]  
        L_{\text{DS}}(G,D,M) = L_{\text{GAN}}(G,D) + \beta L_{\text{con}}(G,M),
    \end{aligned}
\label{eqn3}
\end{equation}
where $L_{con}(G,D,M)$ is the loss for constraints that encourages the domain shifter to be differentiated, $M$ denotes possibly existing additional modules for the constraint loss, and $\beta$ is a weight that balances the two losses.
Practical implementation details for diversifying domain shifters will be introduced in section~\ref{sec:ImpleDomShifter}.

\subsection{Multi-domain-invariant Representation Learning}
In conventional pixel-level adaptation methods, substantial training of the inference layer heavily depends on the translated source images.
However, these methods run the risk of imperfect image translation, which can cause another domain shift issue with the target domain.
To address this limitation, we design an adversarial learning scheme called Multi-domain-invariant Representation Learning (MRL), which encourages domain-invariant features among the diversely scattered domains through adversarial learning.
We assume that we have $(n+2)$ number of diversified domains with a pairwise domain gap. 
For instance, we regard the translated source domain as separate from the source or the target domain and consider the three domains for conventional pixel-level adaptation methods. 
Most existing feature-level adaptation methods apply adversarial learning through the binary domain discriminator.
However, these domains have pairwise domain shifts given by the domain adaptation problem or caused by the imperfect image translation.
Thus, regarding multiple domains as the same domain during adversarial learning can fatally disturb the model from learning common features.
Thus, we use the discriminator with $(n+2)$ outputs so as to learn to distinguish the domains using the cross entropy loss.

Adversarial learning methods attain domain-invariant features by inducing a feature which confuses the domain discriminator.
This confusion can be achieved by designating each domain to resemble the other in cross-domain adaptation problems.
However, in a multi-domain situation, it is not desirable to specify each domain to resemble each specific target domain.
To address this issue, inspired by \cite{DeepAlign2_GRL}, we attach a gradient reverse layer (GRL) at the front-end of the discriminator.
Since the GRL forces the generator to manufacture the features of the given images as if they were not sampled from its domain, the features of each domain are encouraged to be domain-invariant.
The objective function for MRL can be written as:
\begin{equation}
    \begin{aligned}[b]    
        L_{\text{mrl}}(x^{f}, D_{x^{f}})&=-\sum_{i=0}^{n+1}\sum_{u,v} {\bf 1}_{\{i\}}(D_{x^{f}})log(p_{i}^{\text{(u,v)}}(x^{f}))
    \end{aligned}
\label{eqn4}
\end{equation}
where $x^{f}$ is the feature map given for the discriminator, $1_{\{i\}}$ is the indicator function for a singleton $\{i\}$, $p_{i}^{(u,v)}$ is the domain probability for the $i$th domain of the feature vector located at (u, v) of $x^{f}$, and $D_{x^{f}}$ is the ground-truth for the domain label of $x^{f}$.

\subsection{Structured Domain Adaptation framework for Object Detection}
In this section, we structurize our learning paradigm by integrating DD and MRL into a framework.
Without loss of generality, we assume that there is $n$ number of domain shifters $G_i$ for $i=1, ..., n$.
Our framework aims to learn domain-invariant representation and adapt the object detector for these representations simultaneously.
To achieve the goal, every $(n+2)$ number of domains is utilized for MRL, while the source domain and the shifted domains encourage the localization layers and the classification layers of the object detector. 
The objective function for the framework can be written as follows:
\begin{equation}
    \begin{aligned}[b]    
\mathcal{L}(x^{s}, x^{t}, y^{s}) &= L_{\text{MRL}}( x^{s}, x^{t}) + L_{\text{LOC}}( x^{s}, y^{s}) \\
&+ L_{\text{CLS}}( x^{s}, y^{s}),
    \end{aligned}
\label{eqn5}
\end{equation}
\begin{equation}
    \begin{aligned}[b]   
L_{\text{MRL}}( x^{s}, x^{t})&=L_{\text{mrl}}(G_{\text{Base}}(x^{s}),0)\\
&+ L_{\text{mrl}}(G_{\text{Base}}(x^{t}), n+1)\\
&+ \sum_{i=1}^{n}L_{\text{mrl}}(G_{\text{Base}}(G_i(x^{s})), i),
    \end{aligned}
\label{eqn6}
\end{equation}
\begin{equation}
L_{\text{LOC}}( x^{s}, y^{s})=L_{\text{loc}}( x^{s}, y^{s}) + \sum_{i=1}^{n} L_{\text{loc}}( G_i(x^{s}), y^{s}),
\end{equation}
\begin{equation}
L_{\text{CLS}}( x^{s}, y^{s})=L_{\text{cls}}( x^{s}, y^{s}) + \sum_{i=1}^{n} L_{\text{cls}}( G_i(x^{s}), y^{s}),
\end{equation}
Here, $x^s$ and $x^t$ are images of the source and the target domain, $G_\text{Base}$ is the base convolutional network in Fig.~\ref{fig:NetArch} and $y^{s}$ is the label information for $x^s$.
In addition, $L_{\text{loc}}$ and $L_{\text{cls}}$ denote the regression loss and classification loss for the given image, respectively.
The overall framework is shown in Fig.~\ref{fig:NetArch}.


\section{Experiments}

\subsection{Datasets}
We verify the effectiveness of our learning paradigm in two different settings: 1) adaptation from real-world to artistic media; 2) adaptation among urban scenes.\\

\noindent{\bf Real-world Dataset.}
PASCAL VOC~\cite{PascalVoc} is a real-world image dataset used for several computer vision tasks.
PASCAL VOC 2007 dataset consists of 2,501 train images, 2,510 validation images, and 4,952 test images, while PASCAL VOC 2012 dataset contains 5,717 train images and 5,823 validation images.
Annotations are provided for 20 categories.
We use train set and validation set on PASCAL VOC 2007 and train set and validation set on PASCAL VOC 2012 as a real-world dataset.\\

\noindent{\bf Artistic Media Datasets (AMDs).} We use Clipart1k, Watercolor2k, and Comic2k~\cite{Inoue_2018_CVPR} for artistic media domains.
These datasets are collected from a website called Behance for the image classification task by ~\cite{BAM}.
Recently, Inoue et al.~\cite{Inoue_2018_CVPR} notated labels for the object detection task.
Each dataset consists of 1,000, 2,000, and 2,000 images, respectively, while half of them are for the test set.\\

\noindent{\bf Urban Street Datasets (USDs).}
We use Cityscapes~\cite{CityscapesDataset} and Foggy Cityscapes~\cite{Foggy} for urban scene datasets.
Both of them consist of 2,975 train images and 500 validation images with 8 categories.\\

\noindent{\bf Experiment Setup.}
To validate our method for adaptation tasks from real-world to artistic media, we conduct experiments for Real-world$\rightarrow$Clipart1k, Real-world$\rightarrow$Watercolor2k, and Real-world$\rightarrow$Comic2k.
Whole images of each AMD are used for the target domain data during training, while each test set is used for evaluation.
For urban scenes, we conduct the experiment for Cityscapes$\rightarrow$Foggy Cityscapes. We use Cityscapes train set and Foggy Cityscapes validation set.

\begin{figure*}
\begin{center}
\includegraphics[width=1.0\linewidth]{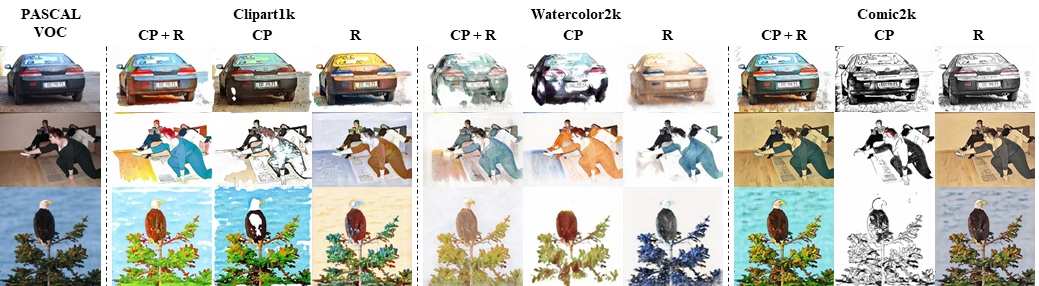}
\end{center}
  \caption{Qualitative results for the shifted domains with various configurations of constraint factors. CP and R denote color preservation constraint and reconstruction constraint, respectively.}
\label{fig:DomDivFig}
\end{figure*}
\subsection{Implementation Details for Domain Shifters}  \label{sec:ImpleDomShifter}
To verify the effectiveness of DD, we generated 3 distinct shifted domains for each adaptation task.
Under the universality for domain shifter architecture, we adopt the residual generator and the discriminator from CycleGAN~\cite{CycleGAN}.
To distinctively shift the source domain, we consider two factors in the objective function, i.e., color preservation and reconstruction.
Figure~\ref{fig:DomDivFig} shows the visual differences caused by each configuration of the factors.\\

\noindent{\bf Domain shift considering color preservation:} 
To constraint the domain shifter to preserve color, we adopt the $L^1$ loss between an input image and a translated image.
However, since the instability of the training increases as we give the less effective constraint, we only assign the constraint to the target domain for the diverse shift.
Thus, the constraint loss for the domain shifter can be written as:
\begin{equation}
L_{\text{con,1}}(G) = \mathbb{E}_{x\sim p_{t}(x)}[\lVert(G(x)-x)\rVert_1].
\end{equation}




\noindent{\bf Domain shift considering reconstruction:} To consider the reconstruction, we need one more pair of domain shifter $G'$ and discriminator $D'$ for inverse translation.
Moreover, we need additional generative adversarial losses necessary for training $G'$.
Thus, the constraint loss for the domain shifter can be written as:

\begin{equation}
\begin{aligned}
L_{\text{con,2}}(G,G',D') &=  \mathbb{E}_{x\sim p_{\text{s}}(x^s)}[logD'(x^s)]\\
&+\mathbb{E}_{x^t\sim p_{\text{t}}(x^t)}[log(1-D'(G'(x)))]\\
&+\mathbb{E}_{x^s\sim p_{s}(x^s)}[\lVert(G'(G(x^s))-x^s)\rVert_1]\\
&+\mathbb{E}_{x^t\sim p_{t}(x^t)}[\lVert(G(G'(x^t))-x^t)\rVert_1].
\end{aligned}
\end{equation}

\noindent{\bf Domain shift considering both reconstruction and color preservation:}
To consider two factors simultaneously, we apply the sum of two constraint loss terms with additional modules $G'$ and $D'$:
\begin{equation}
L_{\text{con,3}}(G,G',D') = L_{\text{con,1}}(G) + L_{\text{con,2}}(G,G',D').
\end{equation}

\subsection{Implementation Details for Object Detection}
In our experiments, we use Faster R-CNN~\cite{FasterR-CNN} as our base object detector with VGG-16~\cite{VGG} pretrained on ImageNet.
Each batch consists of $(n+2)$ images where $n$ is a number of shifted domains. 
We alleviate the memory issue through gradient accumulation. 
We train the network for 80k iterations, 50k iteration with a learning rate of 0.001 and the last 30k iterations with a learning rate of 0.0001.
All implementations are done in PyTorch and on a single GeForce Titan XP GPU.

For PASCAL VOC and AMDs, we resize the images to have a length of 600 pixels as its shorter side.
For USDs, we match the shorter side of the image to be a length of 500 pixels. 
We evaluate mean average precisions (mAP) in the test phase, following the IoU threshold of 0.5 in \cite{Inoue_2018_CVPR} and \cite{Chen_2018_CVPR}.
We follow \cite{FasterR-CNN} for unspecified hyper-parameters.
\subsection{Performance Comparison}  \label{sec:PerformanceComparison}
In this section, we compare our method to the state-of-the-art methods (i.e., Domain Adaptive Faster R-CNN (DAF)~\cite{DAFRCNN} and Domain Transfer (DT) stage of \cite{Inoue_2018_CVPR}).
For our methods, We apply three shifted domains implemented in section~\ref{sec:ImpleDomShifter}.



\begin{figure}[t]
\begin{center}
\centering

  \includegraphics[width=1.0\linewidth]{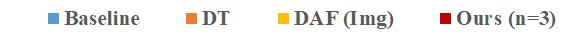}
\vskip -7pt
\subfloat[\footnotesize Watercolor2k]{
\begin{minipage}{0.48\linewidth}
  \centering
      \includegraphics[width=1.0\linewidth]{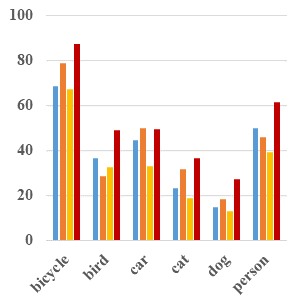}\label{fig:7a}
  \end{minipage}}
    \subfloat[\footnotesize Comic2k]{
  \begin{minipage}{0.48\linewidth}
  \centering
      \includegraphics[width=1.0\linewidth]{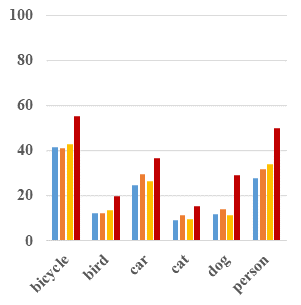}\label{fig:7b}
  \end{minipage} }\label{fig:7b}
  \vskip -3pt
\end{center}
  \caption{Comparison results for the class-wise AP of Watercolor2k test set and Comic2k test set~\cite{Inoue_2018_CVPR}.}
\label{fig:Classwise}
\end{figure}

\begin{figure*}
\begin{center}
  \includegraphics[width=0.98\linewidth]{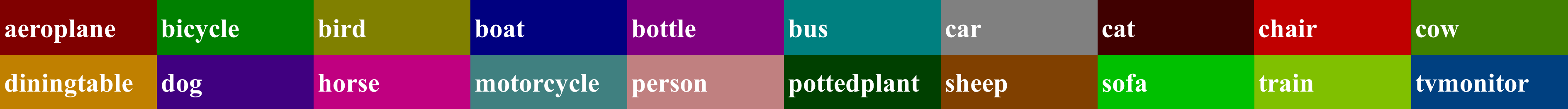}
  \vskip -9pt
  \subfloat[\footnotesize Input image]{
  \begin{minipage}{0.135\linewidth}
  \centering
      \includegraphics[width=1.0\linewidth]{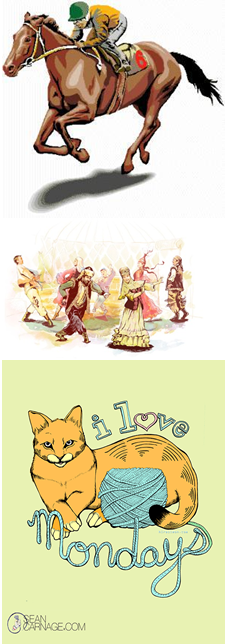}\label{fig:8a}
  \end{minipage}}
  \subfloat[\footnotesize Baseline]{
  \begin{minipage}{0.135\linewidth}
  \centering
      \includegraphics[width=1.0\linewidth]{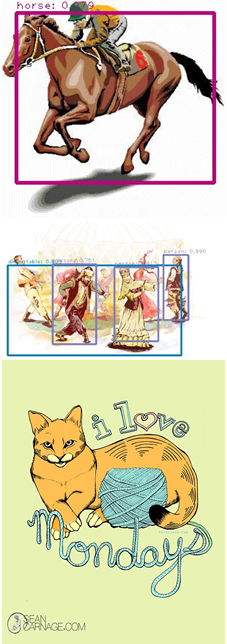}\label{fig:8b}
  \end{minipage}}
  \subfloat[\footnotesize DAF (Img)~\cite{DAFRCNN}]{
  \begin{minipage}{0.135\linewidth}
  \centering
      \includegraphics[width=1.0\linewidth]{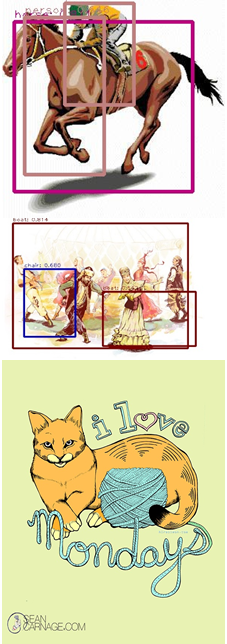}\label{fig:8c}
  \end{minipage}}
  \subfloat[\footnotesize DT~\cite{Inoue_2018_CVPR}]{
  \begin{minipage}{0.135\linewidth}
  \centering
      \includegraphics[width=1.0\linewidth]{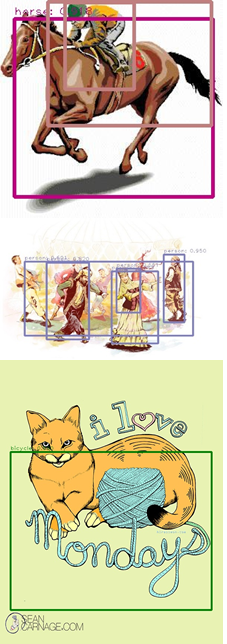}\label{fig:8d}
  \end{minipage}}
  \subfloat[\footnotesize Ours (DD)]{
  \begin{minipage}{0.135\linewidth}
  \centering
      \includegraphics[width=1.0\linewidth]{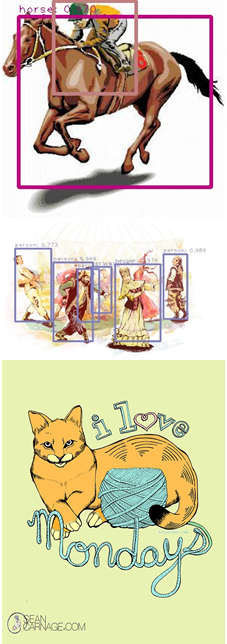}\label{fig:8e}
  \end{minipage}}
  \subfloat[\footnotesize Ours (DD+MRL)]{
  \begin{minipage}{0.135\linewidth}
    \centering
      \includegraphics[width=1.0\linewidth]{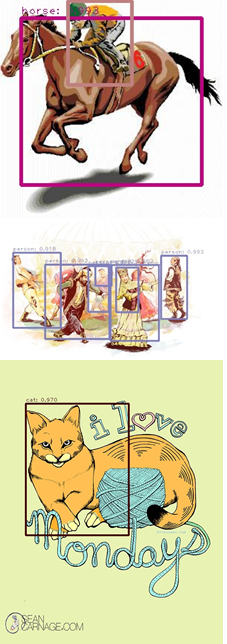}
  \end{minipage}}\label{fig:8f}
  \subfloat[\footnotesize Ground-truth]{
  \begin{minipage}{0.135\linewidth}
  \centering
      \includegraphics[width=1.0\linewidth]{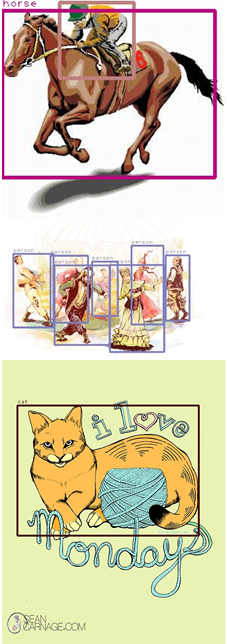}\label{fig:8g}
  \end{minipage}}

\end{center}
  \caption{Qualitative results for object detection of the AMDs by adapting from PASCAL VOC~\cite{PascalVoc}. Images in the first, second, and third rows are from the test sets of Clipart1k, Watercolor2k, and Comic2k~\cite{Inoue_2018_CVPR}, respectively. Best view in color.}
\label{fig:Qualitative}
\end{figure*}

Table~\ref{tab:clip1k},~\ref{tab:WatCom2k},~\ref{tab:foggy}, and Fig.~\ref{fig:Classwise} present the comparison results on Faster R-CNN backbone.
Our learning paradigm achieves the highest class-wise AP among all methods in all adaptation tasks except table class in Clipart1k, car class in Watercolor2k. and bus class in Cityscapes.
Specifically, for the animal classes in AMDs, our proposed method obtains significantly higher class-wise performance than other methods.
To interpret the results in detail, we observe that it is hard to train object detectors with the real-world data to infer discriminatively among animal classes in the artistic media data.
However, our learning scheme significantly improves the performance values for the animal classes.
Moreover, our method exceeds the state-of-the-art methods by 3$\%$ $\sim$ 12$\%$ mAP.
Especially for the Real-world $\rightarrow$ AMD tasks, our method outperforms the state-of-the-art methods by around 9$\%$ $\sim$ 12$\%$ mAP. 
These results demonstrate that our method is effective at learning domain-invariant discriminative features and adapting object detection layers to the common feature space, which is further analyzed in section~\ref{sec:DDcls} and~\ref{sec:ErrAn}.
Several qualitative results are shown in Fig.~\ref{fig:Qualitative}.

\begin{table*}
\small
\begin{center}
\begin{tabular}{|@{\hskip3pt}l@{\hskip2pt}|@{\hskip2pt}c@{\hskip3pt}c@{\hskip3pt}c@{\hskip3pt}c@{\hskip3pt}c@{\hskip3pt}c@{\hskip3pt}c@{\hskip3pt}c@{\hskip3pt}c@{\hskip3pt}c@{\hskip3pt}c@{\hskip3pt}c@{\hskip3pt}c@{\hskip3pt}c@{\hskip3pt}c@{\hskip3pt}c@{\hskip3pt}c@{\hskip3pt}c@{\hskip3pt}c@{\hskip3pt}c@{\hskip2pt}|@{\hskip2pt}c@{\hskip2pt}|}
\hline
Method          & aero & bike & bird & boat & bottle & bus & car & cat & chair & cow & table & dog & horse & mbike & person & plant & sheep & sofa & train & tv  &  mAP  \\ \hline \hline
Baseline        & 13.9&	51.5&	20.4&	10.1&	29.5&	35.1&	24.6&	3.0&	34.7&	2.6&	25.7&	13.3&	27.2&	47.9&	37.5&	40.6&	4.6&	9.1&	27.5&	40.2&	24.9

 \\ \hline
DT~\cite{Inoue_2018_CVPR}      
& 16.4 &	62.5 &	22.8 &	31.9 &	44.1 &	36.3 &27.9 &	0.7 &	41.9 &13.1 &	{\bf 37.6} &	5.2 &28.0 &	64.8&	58.2 &42.7 &9.2 &	19.8 &	32.8&	47.3 &	32.1    \\ \hline
DAF (Img)~\cite{Chen_2018_CVPR}& 20.0 & 49.9 & 19.5 & 17.0 & 21.2 & 24.7 & 20.0 & 2.0 & 30.2 & 10.5 & 15.4 & 3.3 & 25.9 & 49.3 & 32.9 & 23.6 & 14.3 & 5.5 & 30.1 & 32.0 & 22.4 \\ \hline
Ours (n=3)      & {\bf 25.8} &	{\bf 63.2} &	{\bf 24.5} &	{\bf 42.4} &	{\bf 47.9} &	{\bf 43.1} &	{\bf 37.5} &	{\bf 9.1} &	{\bf 47.0}	& {\bf 46.7} &	26.8 &	{\bf 24.9} &	{\bf 48.1} &	{\bf 78.7} &	{\bf 63.0} &	{\bf 45.0} &	{\bf 21.3} &	{\bf 36.1} &	{\bf 52.3} & {\bf 53.4} &	{\bf 41.8} \\ \hline
\end{tabular}
\end{center}
\caption{Quantitative results for object detection of Clipart1k \cite{Inoue_2018_CVPR} by adapting from PASCAL VOC \cite{PascalVoc}.}
\label{tab:clip1k}
\end{table*}

\begin{table}
\small
\begin{center}
\begin{tabular}{|l|c|c|}
\hline
Method          & V $\rightarrow$ Wa & V $\rightarrow$ Co  \\ \hline\hline
Baseline  & 39.8 & 21.4 \\ \hline
DT~\cite{Inoue_2018_CVPR}      & 40.0 & 23.5 \\ \hline
DAF (Img)~\cite{Chen_2018_CVPR}&  34.3 & 23.2 \\ \hline
Ours (n=3)      & {\bf 52.0} & {\bf 34.5} \\ \hline
\end{tabular}
\end{center}
\caption{Quantitative results for object detection of Watercolor2k~\cite{Inoue_2018_CVPR} and Comic2k~\cite{Inoue_2018_CVPR} by adapting from PASCAL VOC~\cite{PascalVoc}. We denote PASCAL VOC, Watercolor2k, and Comic2k as V, Wa, and Co, respectively.}
\label{tab:WatCom2k}
\end{table}

\subsection{Ablation Study on Numbers of Shifted Domains}
We investigate the effectiveness of the DD stage and the MRL stage on different numbers of the shifted domains.
We used the Real-world $\rightarrow$ Clipart1k task as a study case.
As shown in Table~\ref{tab:AblationNumber}, the overall results of each learning scheme are improved as the number of shifted domains increases.
Furthermore, using DD with MRL significantly boosts the performance for overall cases.
It is noteworthy that the improvement in performance through MRL is amplified as the number of domains increases.
These results validate our hypothesis that DD enhances the domain adaptation effect of the following feature-level adaptation by alleviating the source-biased discriminativity.

\subsection{Study on Alleviation of the Source-biased Discriminativity} \label{sec:DDcls}
To further verify the alleviation of the source-biased discriminativity by DD, we investigate the localization performance of RPN and the classification accuracy of the Fast R-CNN module on the Faster R-CNN baseline.
To compare the positive impact of the domain adaptation methods on the localization capability, We compute mean Intersection-over-Union (mIoU) of the best overlaps predicted from RPN for each instance.
The classification accuracy is evaluated with the target domain instances.
To evaluate the inference capability of the classification layer in the Fast R-CNN module, we provide the ground-truth value for bounding boxes.
We conduct the experiments for the Real-world$\rightarrow$Clipart1k case.

\begin{table}
\small
\begin{center}
\begin{tabular}{|@{\hskip1pt}l@{\hskip1pt}|@{\hskip1pt}c@{\hskip2pt}c@{\hskip2pt}c@{\hskip2pt}c@{\hskip2pt}c@{\hskip2pt}c@{\hskip1pt}c@{\hskip1pt}c@{\hskip1pt}|@{\hskip1pt}c@{\hskip1pt}|}
\hline
Method          & \footnotesize{person} & \footnotesize{rider}  & \footnotesize{car} & \footnotesize{truck} & \footnotesize{bus}  & \footnotesize{train} & \footnotesize{mcycle} & \footnotesize{bicycle} &  \footnotesize{mAP}  \\ \hline \hline
Baseline  & 17.7 & 24.7 & 27.2 & 12.6 & 14.8 & 9.1    & 14.3 & 23.2 & 17.9 \\ \hline
DT~\cite{Inoue_2018_CVPR}    & 25.4	& 39.3	& 42.4 &	24.9&	40.4&	23.1&	25.9&	30.4&	31.5\\ \hline
{DAF (Img)~\cite{Chen_2018_CVPR}} & 22.9 & 30.7 & 39.0 & 20.1 & 27.5 & 17.7   & 21.4 & 25.9 & 25.7 \\ \hline
DAF (Ins)~\cite{Chen_2018_CVPR} & 23.6 & 30.6 & 38.6 &20.8  & {\bf 40.5} & 12.8   & 17.1 & 26.1 & 26.3 \\ \hline
DAF (Cons)~\cite{Chen_2018_CVPR} &25.0  & 31.0 & 40.5 & 22.1 &35.3  & 20.2   &20.0 & 27.1 & 27.6  \\ \hline
Ours (n=3) & {\bf 30.8} & {\bf 40.5} & {\bf 44.3} & {\bf 27.2} & 38.4 & {\bf 34.5} & {\bf 28.4}& {\bf 32.2} & {\bf 34.6} \\ \hline
\end{tabular}
\end{center}
\caption{Quantitative results for object detection of Foggy Cityscapes \cite{Foggy} by adapting from Cityscapes \cite{CityscapesDataset}.}
\label{tab:foggy}
\end{table}
\begin{table}
\small
\begin{center}
\begin{tabular}{c|ccc|ccc}
\hline
\multicolumn{4}{c|}{DD Configuration}  & DD  & DD+MRL & offset \\ \hline
$\#$SD & CP & R & CP + R & \multicolumn{3}{c}{mAP}   \\ \hline\hline
0 & & & & 24.9 & - &-\\
1 & \checkmark & & & 31.2 & 32.4 & +1.2\\
2 &\checkmark & \checkmark &  & 32.5 & 37.8 & +5.3\\
3 &\checkmark & \checkmark & \checkmark & 33.8  & 41.8 & +8.0\\\hline
\end{tabular}
\end{center}
\caption{Results of the ablation study on configuration of the shifted domains. DD and MRL denote domain diversification and multi-domain-invariant representation learning, respectively. The offset denotes the performance improvement of the object detector through MRL. CP, R, CP+R denote the shifted domains trained with color preservation constraint, reconstruction constraint, and both constraints, respectively, and SD denotes shifted domains.
}
\label{tab:AblationNumber}
\end{table}

As shown in Table~\ref{tab:AccmIoU}, all domain adaptation methods significantly improve the localization capability of RPN than baseline.
However, the domain adaptation methods with DD achieve significanly higher classification accuracy than the methods without DD. 
Moreover, even though both DAF and MRL are in a frame of feature-level adaptation, the classification results of two methods show considerable gap.
These results demonstrate the importance of the discriminative feature when adapting the domains in feature level.
Furthermore, we can confirm our demonstration that feature-level adaptation suffers from the source-biased discriminativity and DD is effective at alleviating this issue.  

\subsection{Error Analysis on Top Ranked Detections} \label{sec:ErrAn}

We analyze detection errors to investigate the positive impact of our method on domain adaptation in details.
We study Real-world$\rightarrow$Clipart1k case for the analysis.
Since the Clipart1k test set only has 500 images, we classify the most confident 1,000 detections for each domain adaptation method.
With reference to ~\cite{ErrorAnalysis}, we categorize the detection results into three groups: correct detection, mislocalization error, and background error. Correct detection denotes correct class with IoU greater than 0.5, mislocalization error denotes correct class with IoU between 0.1 and 0.5, and background error denotes wrong class or correct class with IoU less than 0.1, where IoU denotes Intersection-over-Union.

\begin{table}
\small
\begin{center}
\begin{tabular}{|l|c|c|}
\hline
Method & Acc ($\%$) & mIoU ($\%$)    \\ \hline \hline
Baseline  & 30.6 & 56.5  \\ \hline
DAF (Img)  & 38.0  & 65.9   \\ \hline
Ours (DD) & 50.2 & 66.6 \\ \hline
Ours (DD+MRL) & 52.5  &68.5  \\ \hline
\end{tabular}
\end{center}
\caption{Comparison results for the instance classification accuracy of the Fast R-CNN module and mean IoU of RPN for the test set of Clipart1k~\cite{Inoue_2018_CVPR}. Each adaptation method only uses annotations in PASCAL VOC~\cite{PascalVoc}.}
\label{tab:AccmIoU}
\end{table}

\begin{figure}[t]
\begin{center}
\includegraphics[width=1.0\linewidth]{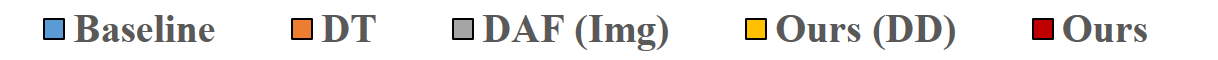}

  \subfloat[\footnotesize Correct]{
\begin{minipage}{0.31\linewidth}
    \centering
      \includegraphics[width=1.0\linewidth]{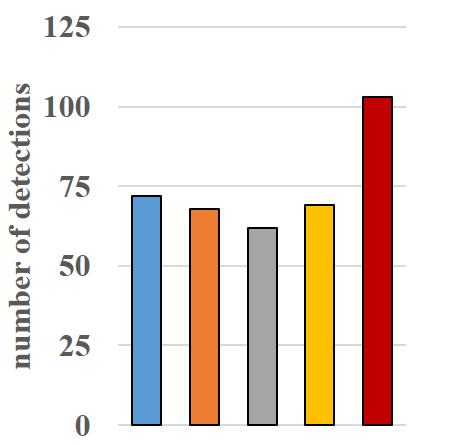}
  \end{minipage}}
  \subfloat[\footnotesize Mislocalization]{
\begin{minipage}{0.31\linewidth}
    \centering
      \includegraphics[width=1.0\linewidth]{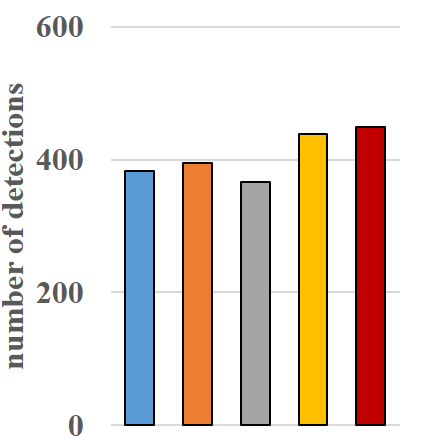}
  \end{minipage}}
  \subfloat[\footnotesize Background]{
\begin{minipage}{0.31\linewidth}
    \centering
      \includegraphics[width=1.0\linewidth]{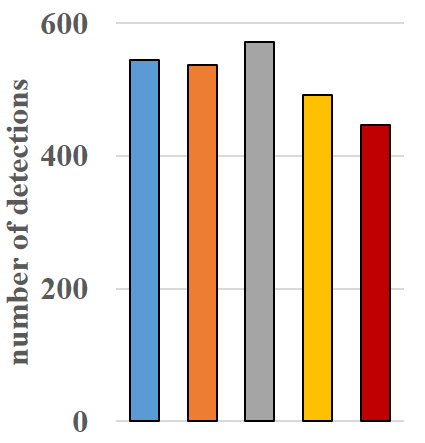}
  \end{minipage}}
   
\end{center}
  \caption{Error Analysis of the most confident 1,000 detections for each domain adaptation methods.}
\label{fig:ErrAn}
\end{figure}

As shown in Fig.~\ref{fig:ErrAn}, both DD with and without MRL reduce background detection errors compared to other methods.
However, while both reduce background errors, DD with MRL significantly increases the number of correct detection than DD.

\section{Conclusion}
In this paper, we have introduced a learning paradigm for object detection to alleviate the chronic limitations of domain adaptation approaches.
Our learning paradigm achieves the goal with the incorporation of Domain Diversification (DD) and Multi-domain-invariant Representation Learning (MRL).
DD mitigates the source-biased discriminativity of feature-level adaptation by diversifying the distribution of the labeled data.
MRL addresses the imperfect image translation by encouraging the unbiased semantic representation among multiple domains.
We structurized our learning paradigm into a domain adaptation framework for object detection networks.
We confirmed the positive impact of DD and MRL through in-depth analysis, which verifies the effectiveness of our two schemes.
Our method outperforms state-of-the-art methods in various cases.\\

\noindent{\bf Acknowledgements} This work was supported by Institute for Information $\&$ communications Technology Promotion(IITP) grant funded by the Korea government(MSIT) (No.2017-0-01772, Development of QA systems for Video Story Understanding to pass the Video Turing Test).

{\small
\bibliographystyle{ieee}
\bibliography{egbib}
}

\end{document}